\documentclass[10pt,twocolumn,letterpaper]{article}

\usepackage{cvpr}              

\usepackage{graphicx}
\usepackage{amsmath}
\usepackage{amssymb}
\usepackage{booktabs}
\usepackage{multirow}
\usepackage[accsupp]{axessibility}  
\usepackage[pagebackref,breaklinks,colorlinks]{hyperref}

\usepackage[capitalize]{cleveref}
\crefname{section}{Sec.}{Secs.}
\Crefname{section}{Section}{Sections}
\Crefname{table}{Table}{Tables}
\crefname{table}{Tab.}{Tabs.}

\def\cvprPaperID{121} 
\def\confName{CVPR}
\def\confYear{2022}

\begin{document}

\title{MANIQA: Multi-dimension Attention Network for  No-Reference \\Image Quality Assessment}

\author{Sidi Yang\thanks{Contribute equally. This work was performed when Tianhe Wu was visiting Tsinghua University as a research intern.}\quad Tianhe Wu\footnotemark[1]\quad Shuwei Shi\quad Shanshan Lao\quad Yuan Gong\quad \\Mingdeng Cao\quad Jiahao Wang\quad Yujiu Yang\thanks{Corresponding author.}\\
 Tsinghua Shenzhen International Graduate School, Tsinghua University\\
{\tt\small \{yangsd21,\ ssw20,\ laoss21,\ gang-y21,\ cmd19,\ wang-jh19\}@mails.tsinghua.edu.cn}\\
{\tt\small tianhe\_wu@foxmail.com, yang.yujiu@sz.tsinghua.edu.cn}
}
\maketitle

\begin{abstract}
No-Reference Image Quality Assessment (NR-IQA) aims to assess the perceptual quality of images in accordance with human subjective perception. 
Unfortunately, existing NR-IQA methods are far from meeting the needs of predicting accurate quality scores on GAN-based distortion images. 
To this end, we propose \textbf{M}ulti-dimension \textbf{A}ttention \textbf{N}etwork for no-reference \textbf{I}mage \textbf{Q}uality \textbf{A}ssessment (MANIQA) to improve the performance on GAN-based distortion. 
We firstly extract features via ViT, then to strengthen global and local interactions, we propose the Transposed Attention Block (TAB) and the Scale Swin Transformer Block (SSTB). 
These two modules apply attention mechanisms across the channel and spatial dimension, respectively. In this multi-dimensional manner, the modules cooperatively increase the interaction among different regions of images globally and locally.
Finally, a dual branch structure for patch-weighted quality prediction is applied to predict the final score depending on the weight of each patch’s score. 
Experimental results demonstrate that MANIQA outperforms state-of-the-art methods on four standard datasets (LIVE, TID2013, CSIQ, and KADID-10K) by a large margin. 
Besides, our method ranked first place in the final testing phase of the NTIRE 2022 Perceptual Image Quality Assessment Challenge Track 2: No-Reference.
Codes and models are available at  \href{https://github.com/IIGROUP/MANIQA}{https://github.com/IIGROUP/MANIQA}.

\end{abstract}

\begin{figure}[ht]
    \centering
    \includegraphics[width=1.0\linewidth]{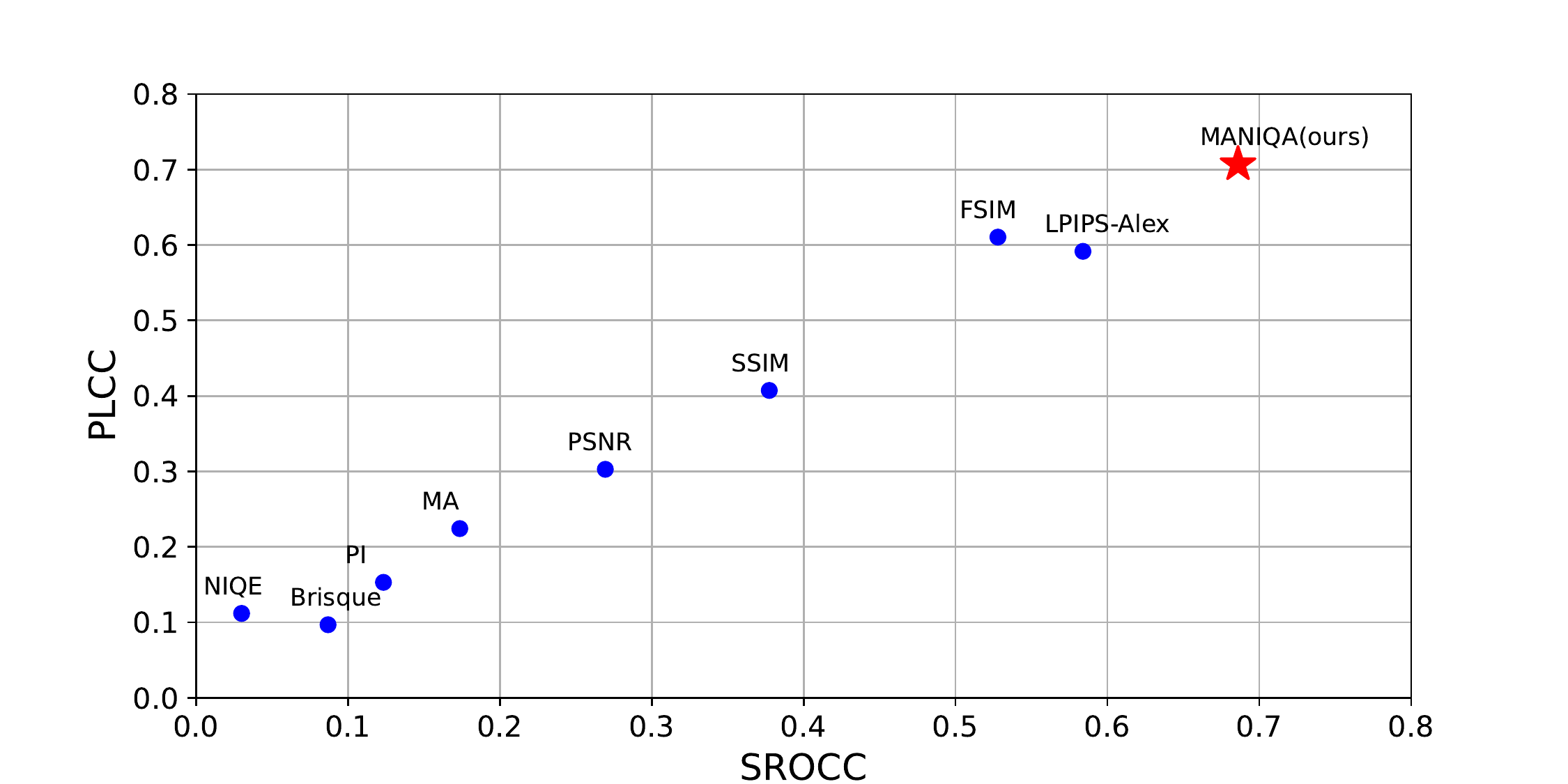}
    \caption{SROCC v.s PLCC results of different methods for NR-IQA on validation of NTIRE 2022.}
    \label{fig:my_label}
\end{figure}
\section{Introduction}
\label{sec:intro}
Pristine-quality images could be degraded in any stage of the whole circulation, from acquisition, compression, storage, transmission\cite{zhu2020metaiqa} to restoration by GAN-based algorithms\cite{goodfellow2014generative, wang2018esrgan}, which results in the loss of received visual information. 
These low-quality (LQ) images have adverse impacts on society. 
On the one hand, billions of photos are uploaded and shared on social media platforms such as Facebook,  Instagram, \emph{etc}, which means that LQ images could badly affect the user's visual feelings. 
On the other hand, for the sake of security, LQ images sometimes impede the normal running of autonomous vehicles\cite{lou2019veri, chiu2020assessing}. 
For these reasons, predicting the perceptual image quality accurately is crucial for industries and daily life.\cite{golestaneh2022no}. 

In general, objective quality metrics can be divided into full-reference (FR-IQA) and no-reference (NR-IQA) methods, according to the availability of reference images\cite{wang2006modern}. 
Previous general-purpose NR-IQA methods focus on quality assessment for distorted synthetic (\eg, Blur, JPEG, Noise) and authentic images (\eg, poor lighting conditions, sensor limitations)\cite{saad2012blind, mittal2012no, zhang2015feature}. 
However, with the extensive application of GAN-based image restoration (IR) algorithms, although these methods can maintain consistency with human subjective evaluation in synthetic or authentic datasets, they show limitations when assessing the images generated by GAN-based IR algorithms\cite{jinjin2020pipal}. In this regard, Gu \etal\cite{jinjin2020pipal} proposed an IQA benchmark including proportioned GAN-based distorted images and previously proposed NR-IQA methods which have shown, as could be expected, unsatisfying performance on the dataset\cite{Gu_2021_CVPR}. 
This gap stems from the inconformity of the Human Visual System and machine vision. 
Users mostly expect rich and realistic details while the machines focus on distinguishing the misalignment of degraded images and pristine-quality images. Specifically, the restored images often possess unreal lifelike textures that satisfy the human eyes' perception but deviate from the prior knowledge learned by deep-learning models for IQA. 

Considering the deficient methods for rating GAN-based distorted images, we propose the Multi-dimension Attention Network for no-reference Image Quality Assessment (MANIQA). 
The proposed method consists of four components: feature extractor using ViT\cite{dosovitskiy2020image}, transposed attention block, scale swin transformer block, and a dual branch structure for patch-weighted quality prediction. 
We first extract and concatenate  4 layers' features from ViT, then compute the weights of different channels through the proposed Transposed Attention Block (TAB). 
We apply self-attention (SA) algorithm across channel rather than the spatial dimension to compute cross-covariance across channels to generate an attention map in this module. Two advantages are provided by this module. First, 4 layers' features from ViT contain different information in channels about the input image. TAB rearranges channels' weight in terms of their importance to perceptual quality score. Second, the attention map generated by this block encodes the global context
implicitly\cite{zamir2021restormer}, which is complementary to downstream local interaction.
To strengthen the local interaction among patches of an image, the Scale Swin Transformer Block (SSTB) is applied. For stabilizing the training process, a scale factor $\alpha$ is applied to adjust the residuals. 
These two modules (TAB and SSTB) apply attention mechanisms across the channel and spatial dimension, respectively. In this multi-dimensional manner, the modules cooperatively increase the interaction among different regions of images globally and locally.
Finally, a dual branch structure consists of weighting and scoring branches for each patch's importance and quality prediction is proposed to attain the final score of images. We assume the salient subjects which mostly in the center of images are noticeable for Human Visual System but not always high-quality. 
On account of the inconsistency of notable and high-
quality regions, the final weighted maps balanced the difference of both via Hadamard Product.

To verify our method's generalization, we have conducted experiments on the PIPAL and other four datasets (TID2013, LIVE, CSIQ, and KADID-10K). 
The results reveal the impressive performance of our proposed method. 
Our method outperforms state-of-the-art(SoTA) on all the above datasets by a large margin. 
Besides, our ensemble method ranked first place in the NTIRE 2022 Perceptual Image Quality Assessment Challenge Track 2: No-Reference (NTIRE 2022 NR-IQA Challenge) which indicates the effectiveness of handling GAN-based distortion of our model.
\begin{figure*}[ht]
    \centering
    \includegraphics[width=\linewidth]{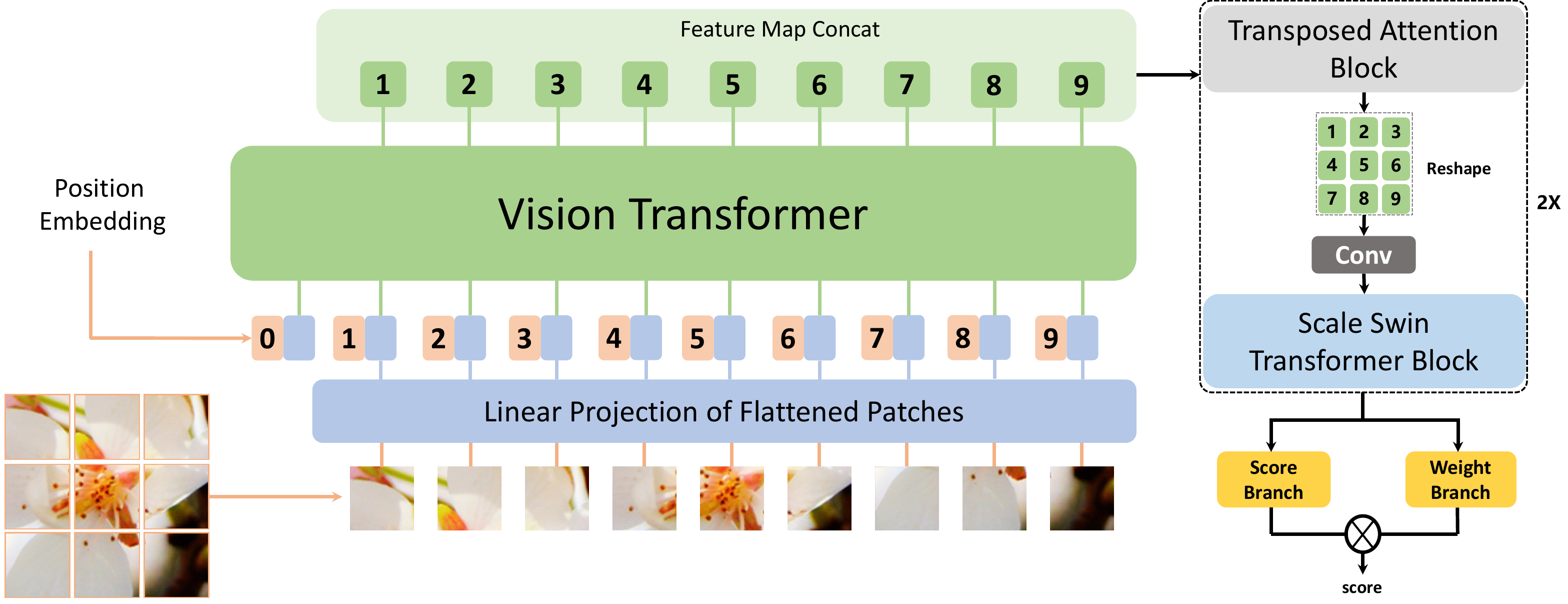}
    \caption{The architecture of the proposed approach - Multi-dimension Attention Network for no-reference Image Quality Assessment(MANIQA). 
    A distorted image is cropped into $8\times8$ sized patches. 
    Then the patches are inputted into the Vision Transformer (ViT) for extracting the features. Transposed attention block and scale swin transformer block, which are described in detail in Sec.~\ref{sec:CA} and Sec.~\ref{sec:swin}, are used to strengthen the global and local interaction. 
    A dual branch structure is proposed for predicting the weight and score of each patch in Sec.~\ref{patch}.}
    \label{fig:overall}
\end{figure*}

\section{Related Work}
\label{sec:Rel_Work}
\noindent\textbf{No-Reference Image Quality Assessment.} Unlike the full-reference (FR-IQA) methods, no-reference (NR-IQA) methods can only use low-quality (LQ) images as input to measure image quality without any reference directly. 
Previous general-purpose NR-IQA methods are mainly divided into natural scene statistics (NSS) based metrics \cite{saad2012blind,mittal2012making,zhang2015feature,moorthy2010two,moorthy2011blind,xu2016blind,ghadiyaram2017perceptual,gao2013universal} and learning-based metrics \cite{ye2012no,ye2012unsupervised, ye2014beyond, zhang2014training, zhang2015som, ma2017dipiq}. 
Based on the assumption for these hand-crafted feature-based approaches
that the natural scene statistics (NSS) extracted from natural images are highly regular, the domain variation caused by synthetic distortions such as spatial\cite{mittal2012no, zhang2015feature, mittal2012making}, gradient\cite{mittal2012no}, discrete cosine transform (DCT)\cite{saad2012blind} and wavelet\cite{moorthy2011blind}, are detectable by these traditional methods. 
In contrast to synthetic distortions in which degradation processes are precisely specified and can be simulated in laboratory environments, authentic distortions collected in the wild have more types of complex distortions. 
However, the hand-crafted feature-based approaches performing well on type-specific distorted datasets have less ability to model real-world distortions. 
With the success of deep learning and Convolution Neural Network (CNN) in many computer vision tasks, CNN-based methods have significantly outperformed previous approaches in terms of real-world distortions by directly extracting discriminative features from LQ images\cite{kang2014convolutional, bosse2017deep, lin2018hallucinated, talebi2018nima, ma2017end, zhang2018blind, bianco2018use, zhu2020metaiqa, su2020blindly, xia2020domain}. 
\cite{kang2014convolutional, bosse2017deep, lin2018hallucinated, zhu2020metaiqa, su2020blindly} leveraged the extracted features to predict the subjective score with the help of pre-trained CNN-based models. 
Hyper-IQA\cite{su2020blindly} separated the features into the low-level and high-level features and transformed the latter to redirect the former. 
Meta-IQA\cite{zhu2020metaiqa} used meta-learning to train the network with separate distortion types for learning prior knowledge. 
Hallucinated-IQA \cite{lin2018hallucinated} exploited the advantages of generative adversarial models. The model estimates the perceptual quality of images with reference images generated by a generative network. 

\vspace*{3pt} 
\noindent \textbf{IQA for GAN-based IR Algorithm.} In recent years, GAN\cite{goodfellow2014generative} has been widely used for restoring LQ images (\eg, deblur, denoise, super-resolution) and some works focus on measuring image restoration algorithms using IQA methods\cite{zheng2021learning, blau2018perception}. 
On account of the mechanism of GAN, some unreal textures are lifelike, which are hard to distinguish by human eyes but easy for machines to perceive\cite{qian2020thinking}. 
Gu \etal \cite{jinjin2020pipal} tested several existing algorithms and found algorithms’ tolerance toward spatial misalignment may be a key factor for their performances when assessing GAN-based images. 
In this light, we proposed a new model to handle the GAN-based distorted images in PIPAL.

\noindent \textbf{Transformer for IQA.} In recent years, CNNs have been the main backbone in most computer vision tasks, including FR-IQA and NR-IQA. Unfortunately, the internal disability of capturing non-local features and strong locality bias hinder the models to exploit the information across all regions of an image~\cite{golestaneh2022no}. 
Besides, the spatial translation invariance of shared convolution kernel weights makes CNNs powerless to handle complex combinations of features. 
Inspired by NLP, which applies Transformers\cite{vaswani2017attention} to capture the global dependencies of features, great promotions in various computer vision tasks following the success of vision transformer(ViT)\cite{dosovitskiy2020image} have been achieved. 
In IQA, IQT \cite{you2021transformer} leveraged the reference and distortion image features extracted by CNNs as the input of Transformer for the quality prediction task. 
MUSIQ \cite{ke2021musiq} used Transformer to encode 3 scales of distortion image features and settled the incompatibility of different input image sizes during training and testing. TReS applied relative ranking and self-consistency loss to utilize the rich self-supervisory information and reduce the sensitivity of the network. 
Different from these methods, we delve into the multi-dimension feature interaction and leverage the spatial and channel structure information to compute a non-local representation of the image. The results reveal that our method achieves best performance in five datasets.

\vspace*{2pt} 
\noindent \textbf{Attention Mechanism.} Attention mechanisms have been
widely used in various computer vision tasks \cite{liu2017robust, vaswani2017attention, woo2018cbam, zamir2021restormer, cao2022vdtr}. 
In NR-IQA, Yang \etal \cite{yang2019sgdnet} proposed an end-to-end
saliency-guided architecture and applied spatial and transposed attention to their model. 
Restormer \cite{zamir2021restormer} applied attention mechanism in a new manner which interchanges the position of spatial and channel dimensions to reduce computation and implicitly fuse the global features. 
Inspired by this work, we proposed transposed attention block to capture the dependencies of features
to attain weighted channel features of the input image.

\section{Proposed Method}
\label{sec:method}
\begin{figure}[ht]
    \centering
    \includegraphics[width=\linewidth]{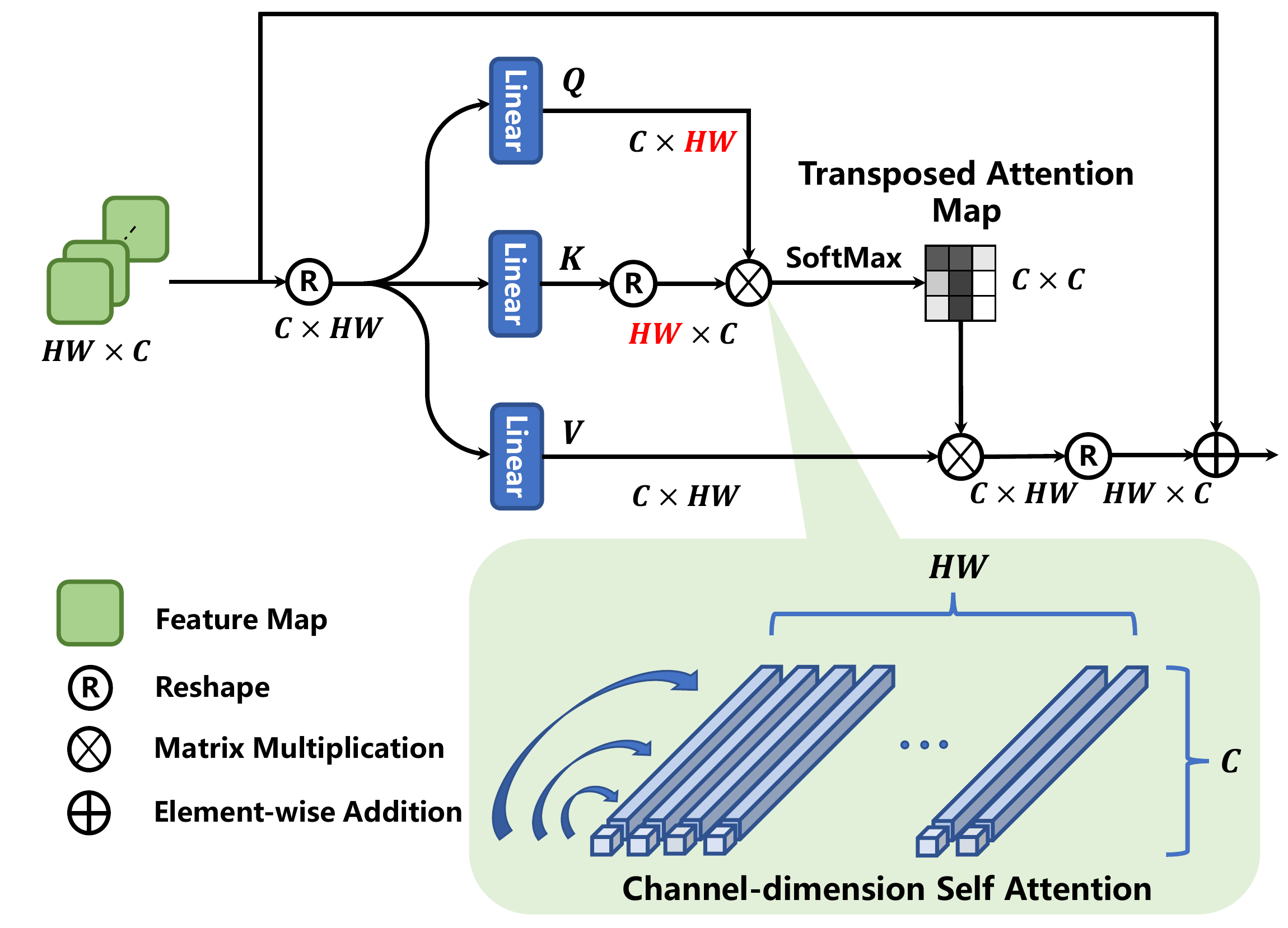}
    \caption{Transposed Attention Block.}
    \label{fig:TA}
\end{figure}

Our main goal is to develop a model that can handle multi-dimension information from the extracted image features. 
To use the information comprehensively from spatial and channel dimensions, we first present the overall pipeline of our MANIQA structure (see in Fig.~\ref{fig:overall}). 
Then we describe the core components of the proposed model: \textbf{(a)} Transposed Attention Block(TAB), \textbf{(b)} Scale Swin Transformer Block(SSTB) and \textbf{(c)} a dual branch structure for patch-weighted quality prediction. 
\subsection{Overall Pipeline}
\label{sec:Feature}
Given a distorted image $\boldsymbol{I} \in \mathbb{R}^{H \times W \times 3}$, where $H$ and $W$ denote height and width, our goal is to estimate its perceptual quality score. 
Let $f_\phi$ present the vision transformer(ViT)\cite{dosovitskiy2020image} with learnable parameters $\phi$, and $F_i \in \mathbb{R}^{b \times c_i \times H_i W_i} $ denotes the features from the $i^{th}$ layer of ViT, where $i \in \{1, 2,\cdots, 12\}$, $b$ denotes the batch size, and $c_i$, $H_i$, and $W_i$ denote the channel size, width, and height of the $i^{th}$ feature, respectively. 
We use 4 from the total 12 layers to extract features from different semantic degrees. 
Next we concatenate $\hat{F}_i$, where $i \in \{7, 8, 9, 10\}$, and denote the output by $\Tilde{F} \in \mathbb{R}^{b \times \sum_i c_i \times H_i W_i}$.

Next, the transposed attention block is employed to boost the channel interaction among the extracted features. 
Then the feature $\Tilde{F}$ will be sent to the scale swin transformer block to 
strengthen the interaction of the local information. 
More details of the two modules can refer to Sec.~\ref{sec:CA} and Sec.~\ref{sec:swin}. 
The final feature map will be sent to the dual branch prediction module with the scoring branch and weighting branch. 
The final score can be computed by the multiplication of scores and weights of total patches as:

\begin{equation}\label{q}
    \Tilde{q}=\frac{\sum_{0 < i < N}\omega_i \times s_i}{\sum_{0 < i < N}\omega_i},
\end{equation}
where N denotes the number of patches for one image.

\subsection{Transposed Attention Block}
\label{sec:CA}

Self-attention (SA) layer is vital for Transformer\cite{vaswani2017attention}. 
In conventional SA\cite{dosovitskiy2020image}, the key-query dot-production interaction builds the global connection among patches in the spatial dimension. However, it ignores the rich information among different channels. 
To alleviate this issue, we propose Transposed Attention Block (TAB), shown in Fig.~\ref{fig:TA}, which applies SA across channel rather than the spatial dimension to compute cross-covariance across channels to generate an attention map encoding the global context implicitly.
From the concatenated feature $\Tilde{F} \in \mathbb{R}^{b \times \sum_i c_i \times H_i W_i}$, our TAB first generates \emph{query} (\textbf{Q}), \emph{key} (\textbf{K}) and \emph{value} (\textbf{V}) projections, which are achieved by 3 independent linear projections, to encode the pixel-wise cross-channel context. 
Then, we reshape query and key projections such that their dot-product interaction generates a transposed-attention map $\textbf{A}$ of size $\mathbb{R}^{\Tilde{C} \times \Tilde{C}}$. 
Note that $\Tilde{C}$ is numerically equal to $c_i$. 
We remove the layer normalization and multi-layer perceptron from the original Transformer\cite{vaswani2017attention}. 
Overall, the TAB process is defined as:

\begin{equation}
\begin{aligned}
    \hat{\mathbf{X}} = W_p \operatorname{Attn}&(\hat{\textbf{Q}}, \hat{\textbf{K}}, \hat{\textbf{V}}) + \textbf{X}, \\
    \operatorname{Attn}(\hat{\textbf{Q}}, \hat{\textbf{K}}, \hat{\textbf{V}})&=\hat{\textbf{V}} \cdot \operatorname{Softmax}(\hat{\textbf{K}} \cdot \hat{\textbf{Q}}/\alpha),
\end{aligned}
\end{equation}
where $\alpha$ indicates the spatial dimension of $\mathbf{Q}$, $\mathbf{K}$ and $\mathbf{V}$. 
As the dot-product is conducted across channel dimensions, we remove the layer normalization, which is used for pixel-wise channel-context normalization. 
Multi-layer perceptron in the original transformer block\cite{vaswani2017attention} is also removed in our proposed block as the increased complexity affects the generalization of the model. 
Two advantages are provided by this module. First, 4 layers' features from ViT contain different information in channels about the input image. TAB rearranges channels' weight in terms of their importance to perceptual quality score. Second, the attention map generated by the TAB encodes the global context
implicitly\cite{zamir2021restormer}, which is complementary to downstream local interaction.
\begin{figure}[ht]
    \centering
    \includegraphics[width=\linewidth]{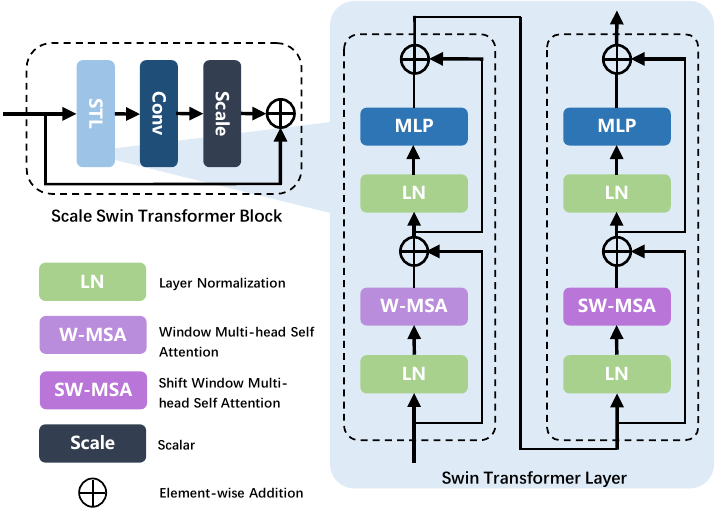}
    \caption{Our Scale Swin Transformer Block.}
    \label{fig:swin}
\end{figure}

\subsection{Scale Swin Transformer Block}
\label{sec:swin}

As shown in Fig.~\ref{fig:swin}, the Scale Swin Transformer Block consists of Swin Transformer Layers (STL)\cite{liu2021swin} and a convolutional layer. 
Given the input feature $F_{i,0}$, the SSTB first encodes the feature through 2 layers of STL:
\begin{figure}
    \centering
    \includegraphics[width=1.0\linewidth]{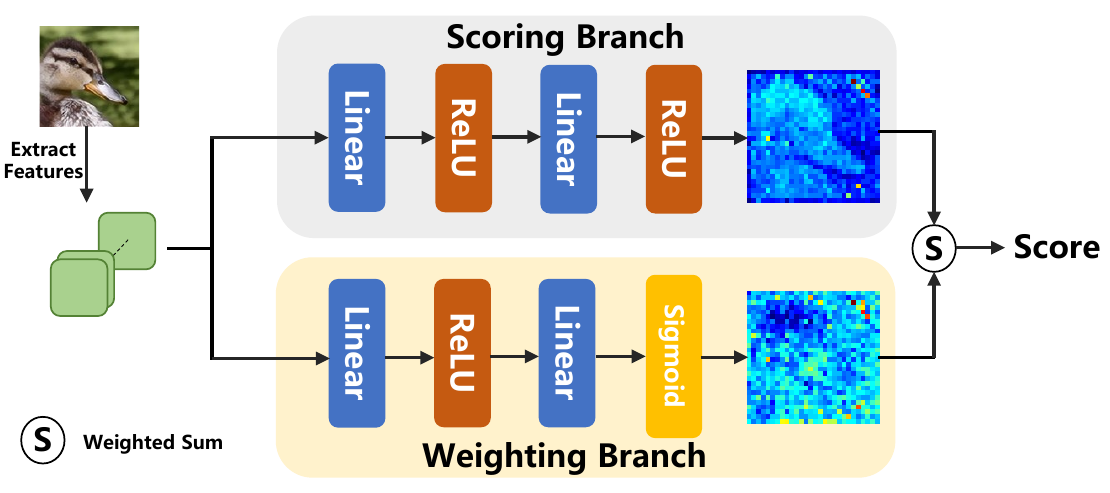}
    \caption{Dual branch structure for patch-weighted quality prediction.}
    \label{fig:patch weight}
\end{figure}
\begin{equation}
    F_{i,j}=H_{STL_{i,j}}(F_{i,j}), \quad j=1,2,
\end{equation}
where $H_{STL_{i,j}}(\cdot)$ is the $j$-th Swin Transformer layer in the $i$-th stage of our block, $i \in \{1,2\}$. 
Then, a convolutional layer is applied before the residual connection. 
The output of SSTB is formulated as:
\begin{equation}
    F_{out}=\operatorname{\alpha}  \cdot H_{CONV}(H_{STL}(\Tilde{F}_{i,2})) + \Tilde{F}_{i,0},
\end{equation}
where $H_{CONV}(\cdot)$ is the Swin Transformer layer. 
$\alpha$ denotes the scale factor of the output of STL. This design has two benefits. 
First, the convolutional layer with spatially invariant filters can enhance the translational equivariance\cite{liang2021swinir}. 
Second, the scale factor $\alpha$ stabilizes the training\cite{szegedy2017inception} through residual connection. 
In general we pick the constant between 0.1 and 0.2 to scale the residuals before adding them to the input.

\subsection{Patch-weighted Quality Prediction}
\label{patch}

We designed a dual branch structure for patch-weighted quality prediction as shown in Fig.~\ref{fig:patch weight}. 
This module consists of a scoring and weighting branch which predicts each patch's score and weight respectively. Given the feature $\Tilde{F}$, we generate $weight$ ($\mathbf{W}$) and $score$ ($\mathbf{S}$) projections, which are achieved by 2 independent linear projections. 
We believe the quality of an image is dependent on its different regions. 
So the final patch score of the distorted image is generated by multiplication of each patch's score and weight, then the final score of the whole image is generated by the summation of final patch scores. In consideration of inhibiting the overfitting of the model, we believe the scoring and weighting branch create a mechanism of restraint. The final score has to take the balance of scoring and weighting branch into account.

\begin{figure*}[ht]
    \centering
    \includegraphics[width=\linewidth]{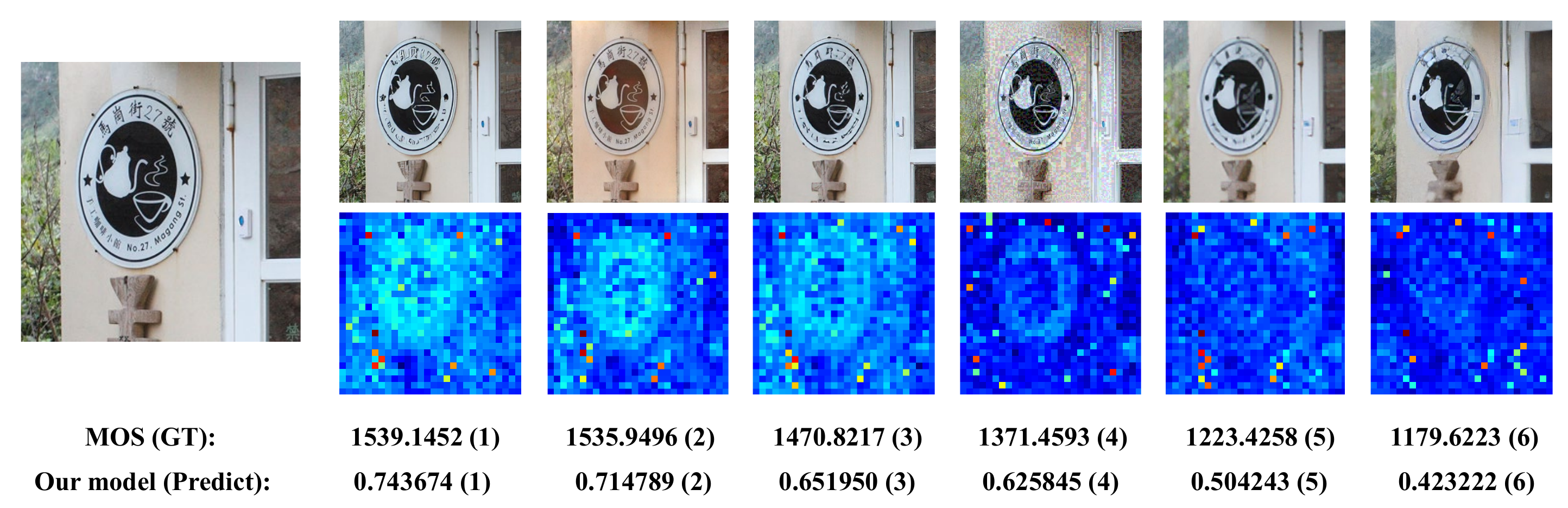}
    \caption{Example images from validation dataset of the NTIRE 2022 challenge. 
    On the left is the pristine-quality image and on the right are distorted images. 
    MOS denotes the ground-truth human rating.  
    The number in the parenthesis denotes the rank among considered distorted images in this figure. 
    The patch weight maps are below the distorted images.}
    \label{fig:vis}
\end{figure*}

\section{Experiments}
\label{sec:exp}

\subsection{Datasets}
PIPAL~\cite{jinjin2020pipal} is a recently proposed IQA dataset, which contains images processed by image restoration and enhancement methods (particularly the deep learning-based methods) besides the traditional distorting methods. 
The dataset contains 29k images in total, including 250 high-quality reference images, each of which has 116 distortions involving more than 1.13 million human judgments. 
It is challenging for existing metrics to predict perceptual quality accurately, especially the images generated by GAN-based IR methods. 
PIPAL is used for both the training and evaluation of the model, and we also conduct the experiments on LIVE\cite{sheikh2006statistical}, CSIQ\cite{larson2010most}, TID2013\cite{ponomarenko2015image} and KADID-10K\cite{lin2019kadid}. 
Tab.~\ref{tab:dataset} shows the summary of the datasets that are used in our experiment.

\begin{table*}[]
    \centering
    \caption{Comparison of MANIQA v.s. state-of-the-art NR-IQA algorithms on three standard datasets. Bold entries in \textbf{\textcolor{black}{black}} and \textbf{\textcolor{blue}{blue}} are the best and second-best performances, respectively. Some data borrowed from \cite{golestaneh2022no}.}
    \label{tab:sota}
    \begin{tabular}{c||cc|cc|cc|cc}
    \hline
    \hline
     &\multicolumn{2}{c|}{LIVE} &\multicolumn{2}{c|}{CSIQ} &\multicolumn{2}{c|}{TID2013} &\multicolumn{2}{c}{KADID-10K} \\
     \cline{2-9}
     &PLCC &SROCC &PLCC &SROCC &PLCC &SROCC &PLCC &SROCC\\
     \cline{1-9}
    DIIVINE\cite{saad2012blind} & 0.908 & 0.892 & 0.776 & 0.804 & 0.567 & 0.643 & 0.435 &0.413 \\
    BRISQUE\cite{mittal2012no} & 0.944 & 0.929 & 0.748 & 0.812 & 0.571 & 0.626 & 0.567 & 0.528 \\
    ILNIQE\cite{zhang2015feature} & 0.906 & 0.902 & 0.865 & 0.822 & 0.648 & 0.521 &0.558 & 0.528 \\
    BIECON\cite{kim2016fully} &0.961 &0.958 &0.823 &0.815 &0.762 &0.717 &0.648 & 0.623 \\
    MEON\cite{ma2017end} & 0.955 &0.951 &0.864 &0.852 &0.824 &0.808 & 0.691 &0.604 \\
    WaDIQaM\cite{bosse2017deep} & 0.955 &0.960 &0.844 &0.852 &0.855 &0.835 &0.752 &0.739 \\
    DBCNN\cite{zhang2018blind} & \textbf{\textcolor{blue}{0.971}} & 0.968 &\textbf{\textcolor{blue}{0.959}} &\textbf{\textcolor{blue}{0.946}} &0.865 &0.816 & 0.856 &0.851 \\
    TIQA\cite{you2021transformer}& 0.965 &0.949 &0.838 &0.825 &0.858 &0.846 &0.855 &0.850 \\
    MetaIQA\cite{zhu2020metaiqa}& 0.959 &0.960 &0.908 &0.899 &0.868 &0.856 &0.775 &0.762 \\
    P2P-BM\cite{ying2020patches}& 0.958 &0.959 &0.902 &0.899 &0.856 &0.862 &0.849 &0.840 \\
    HyperIQA\cite{su2020blindly}& 0.966 &0.962 &0.942 &0.923 &0.858 &0.840 &0.845 & 0.852 \\
    TReS\cite{golestaneh2022no}&0.968 &\textbf{\textcolor{blue}{0.969}} &0.942 &0.922 &\textbf{\textcolor{blue}{0.883}} &\textbf{\textcolor{blue}{0.863}} &\textbf{\textcolor{blue}{0.858}}  &\textbf{\textcolor{blue}{0.915}}\\
    MANIQA(Ours) &\textbf{0.983} &\textbf{0.982} &\textbf{0.968} &\textbf{0.961} &\textbf{0.943}&\textbf{0.937} &\textbf{0.946} &\textbf{0.944}\\
    \hline
    \hline
    \end{tabular}
    
\end{table*}

\begin{table}[]
    \centering
    \caption{Summary of IQA datasets.}
    \label{tab:dataset}
    \begin{tabular}{ccc}
    \hline
    Dataset & \# of Dist. Images & \# of Dist.Types \\
    \hline
    LIVE & 779 & 5 \\
    CSIQ & 866 & 6 \\
    TID2013 & 3,000 & 24 \\
    KADID-10K & 10,125 & 25 \\
    PIPAL & 23,200 & 116 \\
    \hline
    \end{tabular}
    
\end{table}
\subsection{Implementation Details}
\label{imple}
Our experiments are implemented on an NVIDIA GeForce RTX 3090 with PyTorch 1.8.0 and CUDA 11.2 for training and testing. We trained and tested our model on each of the five datasets. Considering the unconformity of input  sizes of ViT and our datasets, we randomly cropped the images into size $224\times 224$ from the training dataset and randomly horizontally flipped the cropped images with a given probability $0.5$.

We choose ViT-B/8\cite{dosovitskiy2020image} as our pre-trained model which is trained on ImageNet-21k and fine-tuned on ImageNet-1k with the patch size $P$ set to $8$. MANIQA contains two stages. Each stage includes $2$ Transposed Attention Block (TAB) and $1$ Scale Swin Transformer Block (SSTB). The first SSTB's embedding dimension $D_1$ is set to $768$ and the second $D_2$ is set to $384$. The hidden layer's dimension of MLP $D_m$, number of heads $H$, and window size is set to $768$, $4$, and $4$ in each SSTB. We set scale $\alpha=0.80$ in SSTB.

Following the standard training strategy from existing IQA algorithms, we randomly split each dataset into $8:2$, $5$ times with selected 5 different seeds. $80\%$ for training and the rest for testing. During training, we set the learning rate $l$ to $1\times 10^{-5}$. The batch size $B$ is set to $8$ and we utilized the ADAM optimizer with weight decay $1\times 10^{-5}$ and cosine annealing learning rate with the parameters $T_{max}$ and $eta_{min}$ set to $50$ and $0$. The training loss we use is Mean Square Error (MSE) Loss. During testing, we randomly cropped $224\times 224$ sized images $20$ times from the original one. The final score is generated by predicting the mean score of these $20$ images and all results are averaged by $10$ times split. We run the experiment $5$ times with different seeds and report the mean metrics for all of the reported results.

\subsection{Evaluation Criteria}
Following the prior works, we use Spearman's rank-order correlation coefficient (SROCC) and Pearson's linear correlation coefficient (PLCC) as the metrics to evaluate the performance of our models.The PLCC is defined as follows:
\begin{equation}
    \operatorname{PLCC} =\frac{\sum^{N}_{i=1}(s_i-\mu_{s_i})(\hat{s}_i-\mu_{\hat{s}_i})}{\sqrt{\sum^{N}_{i=1}(s_i-\mu_{s_i})^{2}}\sqrt{\sum^{N}_{i=1}(\hat{s}_i-\mu_{\hat{s}_i)^{2}}}},
\end{equation}
where $s_i$ and $\hat{s}_i$ respectively indicate the ground-truth and predicted quality scores of $i$-th image, and $\mu_{s_i}$ and  $\mu_{\hat{s}_i}$ indicate the mean of them. $N$ denotes the testing images. Let $d_i$ denote the difference between the ranks of $i$-th test image in ground-truth and predicted quality scores. The SROCC is defined as:
\begin{equation}
    \operatorname{SROCC} =1-\frac{6\sum^{N}_{i=1}d^2_i}{N(N^2-1)}.
\end{equation}
Both metrics, PLCC and SROCC, are in [-1, 1]. A positive value means a positive correlation. Otherwise, the opposite. A higher value indicates better performance.
\subsection{Results}
\noindent \textbf{Evaluation on five datasets.} Tab.~\ref{tab:sota} shows the overall performance comparison on four datasets in terms of PLCC and SROCC. Our method outperforms the existing methods by a significant margin on almost every dataset. Besides, in Tab.~\ref{res_pipal}, the results on PIPAL (also as the dataset of NTIRE 2022 NR-IQA competition) show that our method is powerful when assessing the GAN-based distortion.

\begin{figure}[ht]
    \centering
    \includegraphics[width=\linewidth]{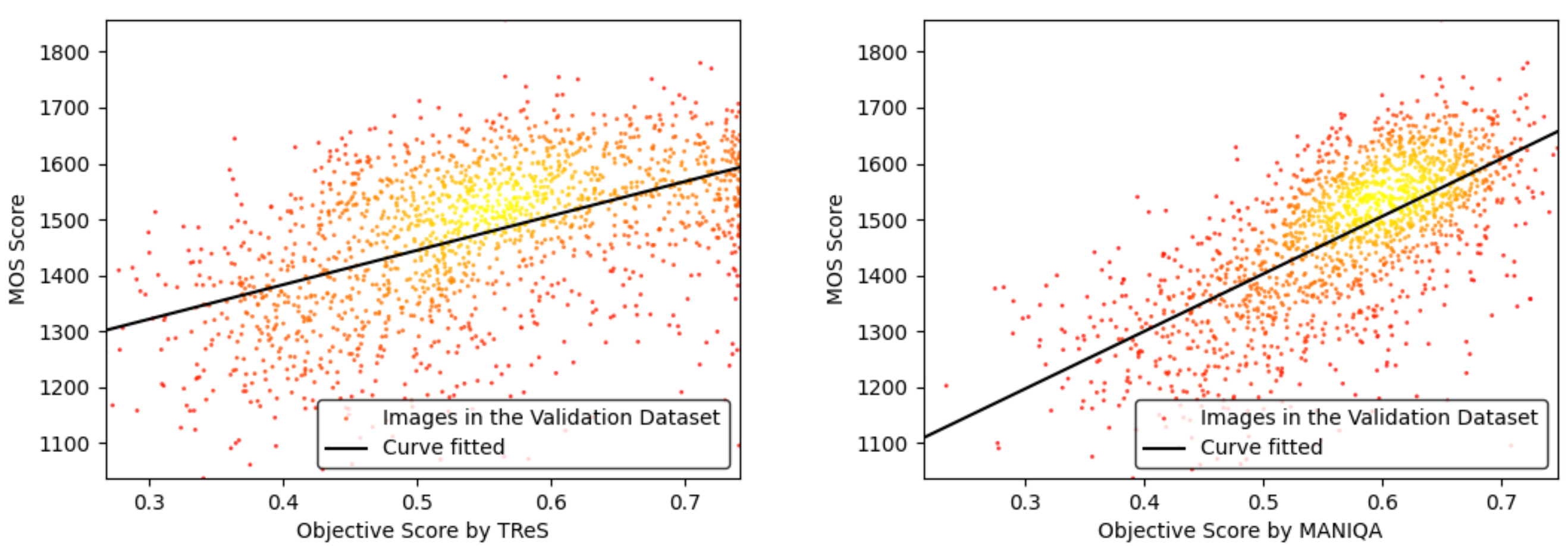}
    \caption{Scatter plots of ground-truth mean opinion scores (MOSs) against predicted scores of TReS and proposed MANIQA on PIPAL datasets.}
    \label{fig:scatter}
\end{figure}

\noindent \textbf{Cross-dataset Evaluation on LIVE and TID2013.} In Tab.~\ref{tab:cross}, we conduct cross dataset evaluations and compare our model to the competing approaches. Training is performed on PIPAL without any finetuning or parameter adaptation. As shown in Tab.~\ref{tab:cross}, our proposed method outperforms the other NR-IQA algorithm and has comparable performance with FR-IQA methods on each dataset, which indicates the strong generalization power of our model. The scatter plot is illustrated in Fig.\ref{fig:scatter}.

\begin{table}[h]
    \centering
        \caption{Performance of different methods on the NTIRE2022 IQA Challenge validation and testing datasets. MANIQA-S and MANIQA-E indicate the single model and ensemble model of MANIQA. Best performances are indicated with \textbf{bold}. Some data borrowed from \cite{gu2022ntire}.}
    \label{res_pipal}
    \scalebox{0.96}{
    \begin{tabular}{c|cc|cc}
    \hline
        \multirow{2}{*}{IQA Name} & \multicolumn{2}{c}{Validation} & \multicolumn{2}{|c}{Test}\\
        \cline{2-5}
         &SROCC &PLCC &SROCC &PLCC \\
         \cline{1-5}
         PSNR &0.250 &0.284 &0.269 &0.303 \\
         SSIM\cite{wang2004image} &0.332 &0.386 &0.377 &0.407 \\
         LPIPS-Alex\cite{zhang2018unreasonable} & 0.581 & 0.616 & 0.584 & 0.592 \\
         FSIM\cite{zhang2011fsim} &0.473 &0.575 &0.528 &0.610 \\
         NIQE\cite{mittal2012making} &0.005 & 0.115 &0.03 & 0.112 \\
         MA\cite{ma2017learning} &0.129 &0.131 &0.173 &0.224 \\
         PI\cite{blau2018perception} &0.079 &0.133 &0.123 &0.153 \\
         Brisque\cite{mittal2011blind} &0.015 &0.059 & 0.087 & 0.097 \\
         MANIQA-S &\textbf{0.686} &\textbf{0.707} &0.667 &0.702 \\
         MANIQA-E &- &- &\textbf{0.704} &\textbf{0.740} \\
         \hline
    \end{tabular}}
\end{table}

\begin{table}
    \centering
    \caption{Evaluations on cross datasets. \emph{FR} refers full-reference IQA methods. \emph{NR} refers no-reference methods. We train each model on PIPAL and test on LIVE and TID2013.}
    \label{tab:cross}
    \scalebox{0.88}{
    \begin{tabular}[width=1.0\linewidth]{c|c|cc|cc}
        \hline
        \multicolumn{2}{c}{Train on} & \multicolumn{4}{|c}{PIPAL} \\
        \hline
        \multicolumn{2}{c|}{\multirow{2}{*}{Test on}} & \multicolumn{2}{c|}{LIVE}  & \multicolumn{2}{c}{TID2013}  \\
        \cline{3-6}
        \multicolumn{2}{c|}{}& PLCC & SROCC & PLCC & SROCC \\
        \hline
        \multirow{3}{*}{FR} &PSNR &0.873 & 0.865 &0.687 &0.677 \\
        \multirow{3}{*}{} &WaDIQaM\cite{bosse2017deep} &0.883 &0.837 &0.698 &0.741 \\
        \multirow{3}{*}{} &RADN\cite{shi2021region} &0.905 & 0.878 &0.747&0.796 \\
        \hline
        \multirow{2}{*}{NR}&TReS\cite{golestaneh2022no} & 0.643& 0.663& 0.516 & 0.563 \\
        &MANIQA(ours) &0.835 &0.855 &0.704 &0.619 \\
        \hline
    \end{tabular}}
\end{table}

\subsection{NTIRE2022 IQA Challenge}

Our method is originally proposed for participating in the NTIRE 2022 Perceptual Image Quality Assessment Challenge Track 2: No-reference. The task of predicting the perceptual quality of an image based on a set of prior examples of images and their perceptual quality labels. For the no-reference track, competitors are only allowed to submit algorithms that take only distorted images as input. 
We use the default settings described in Sec.~\ref{imple}. Besides, the scale factor is set to 0.13. 
Our ensemble approach ranked first place in the final private test phase.

\subsection{Ablation Study}
In Tab.~\ref{tab:abls} and Tab.~\ref{tab:backbone}, we provide ablation experiments to illustrate the effect of different backbones and each component of our proposed method by comparing the results on the PIPAL dataset. Result reported in Tab.~\ref{tab:abls} \#1 gives the baseline performance when none of the components is used.

\vspace*{3pt} 
\noindent \textbf{Feature Extractor.} Tab.~\ref{tab:backbone} shows that ViT serving as the feature extractor has a large edge over the CNN-based network, which means the long-range dependency modeled by transformers is crucial for images. For NR-IQA, pre-trained models trained with small patches are more competitive than the other ones.

\vspace*{3pt} 
\noindent \textbf{Transposed Attention Block.} The transposed attention block is proposed to enhance global interactions among all channels. Results in Tab.~\ref{tab:abls} show such interactions are indispensable for our method. Specifically, this module gains another 0.05 improvements on SROCC and PLCC, respectively when applied with the other two modules. 

\vspace*{3pt} 
\noindent \textbf{Scale Swin Transformer Block.} The scale swin transformer block is proposed to enhance local interactions among all patches. By introducing SSTB, we can observe a clear performance gain in SROCC and PLCC. This demonstrates the effectiveness of local interactions. Besides, In Tab.~\ref{tab:scale}, we test different scale factors on three datasets. Obviously, the model achieves the best performance when the scale factor is set to 0.1.

\vspace*{3pt} 
\noindent \textbf{Patch-weighted Quality Prediction.} A dual branch structure is proposed for estimating the weighted score from each patch. In Tab.~\ref{tab:abls}, this module brings performance gain in SROCC and PLCC. We believe the dual branch struc-
ture could suppress the high weight patches which reduce
the risk of overfitting.

Combined with the above module we proposed, MANIQA significantly improves the evaluation performance.
\begin{table}
    \centering
    \caption{Ablation study on our modules. Results are tested on validation dataset of the NTIRE 2022 IQA Challenge. \emph{TAB} refers to our Transposed Attention Block, \emph{SSTB} refers to our Scale Swin Transformer Block, \emph{Dual Branch} refers to our dual branch structure for patch-weight quality prediction.}
    \label{tab:abls}
    \begin{tabular}{c|ccc|cc}
    \hline
    \# &TAB & SSTB & Dual Branch& SROCC & PLCC \\
    \hline
    1&&&&0.593& 0.607 \\
    2&\checkmark & & &0.598 &0.617 \\
    3&& \checkmark & &0.610 &0.603  \\
    4&\checkmark & & \checkmark& 0.604&0.623  \\
    5& &\checkmark  & \checkmark&0.613 &0.636 \\
    6&\checkmark &\checkmark & \checkmark&\textbf{0.686} &\textbf{0.707} \\
    \hline
    \end{tabular}
    
\end{table}

\begin{table}
    \centering
    \caption{Ablation study of different scale factors on PIPAL dataset.}
    \label{tab:scale}
    \begin{tabular}[width=1.0\textwidth]{c|cc}
    \hline
        Scale & SROCC & PLCC  \\
        \hline
        0.1 &0.686 &0.704\\
        0.2 &0.677 &0.699\\
        0.4 &0.672 &0.696\\
        0.6 &0.669 &0.685\\
        0.8 &0.678 &0.690\\
        1.0 &0.674 &0.689\\
        \hline
    \end{tabular}
\end{table}
\begin{figure}[ht]
    \centering
    \includegraphics[width=\linewidth]{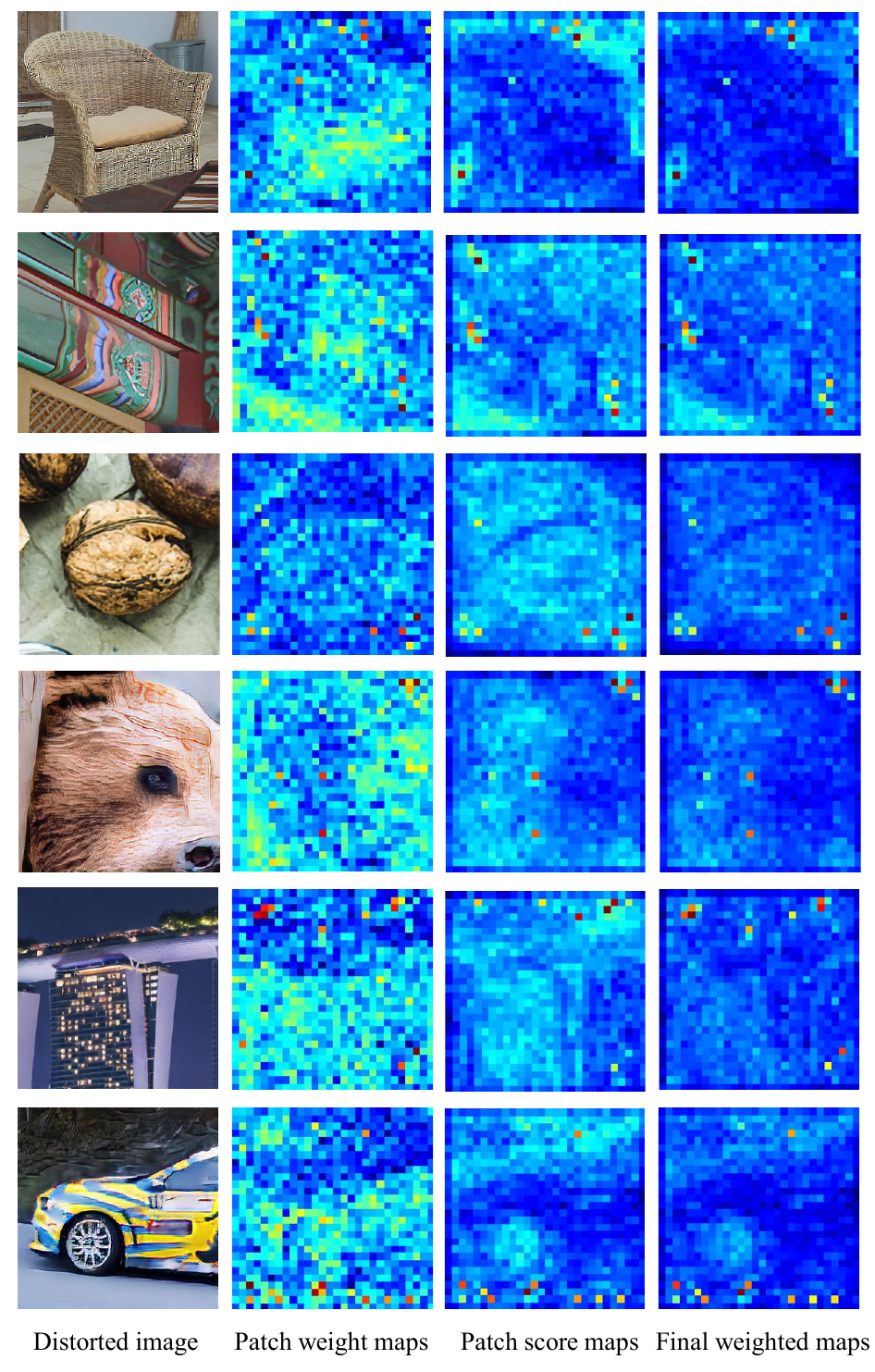}
    \caption{Visualization of weighting branch, scoring branch and final weighted maps of the test dataset of the NTIRE 2022 NR-IQA Challenge. All images are from GAN-based distorted images. Zoom in for more details.}
    \label{fig:vis_5}
\end{figure}
\begin{table}
    \centering
    \caption{Ablation study on the backbone. Results are tested on validation dataset of the NTIRE 2022 IQA Challenge. \emph{Para.} refers to the parameters of models.}
    \label{tab:backbone}
    \begin{tabular}{cc|cc}
    \hline
    Backbone & Para. & SROCC & PLCC \\
    \hline
    ViT-B/8 & 135.62M& 0.686& 0.707 \\
    ResNet50 & 72.98M& 0.512& 0.532 \\
    \hline
    \end{tabular}
    
\end{table}
\subsection{Visualization}
In Fig.~\ref{fig:vis_5}, we visualize the patch weight, score, and final weighted maps of the test dataset of the NTIRE 2022 NR-IQA Challenge generated by our method. As can be seen, the patch weight maps focus on the salient subjects which are noteworthy regions for human eyes, while the patch score maps pay more attention to the regions with a better visual experience. When human observe an image, it is the subject itself that immensely affect our perception. So these patches are given high weights shown in Fig.~\ref{fig:vis_5}. However, the high-quality regions are not always in such regions which contain important semantic information. On account of the inconsistency of notable regions and high-quality regions, the final weighted maps depend on both the patch weight and score maps. We believe the dual branch structure could suppress the high weight patches which reduce the risk of overfitting.

\section{Conclusion}
\label{sec:conclusion}
We proposed Multi-dimension Attention Network for no-reference Image Quality Assessment (MANIQA) and it is appropriately applied to a perceptual image quality assessment task by taking the advantage of the multi-dimension attention network. 
The method has verified the point that multi-dimension interaction in channel and spatial is vital for the perceptual quality assessment task. In the multi-dimensional
manner, the modules cooperatively increase the interaction among different regions of images globally and locally. 
Experiment results on the four standard databases demonstrate outstanding performances compared to the other existing methods.
Besides, our method showed the best performance for GAN-based distortion on PIPAL in the NR track of the NTIRE
2022 Perceptual Image Quality Assessment Challenge. 

Despite the success of our method, there exists room for improvement. In general, not all the GAN-based distortion are annoying. Thus, defining particular distorted types of GAN-based distortions(\eg, pleasing unreal texture, unpleasing unreal texture, \etal) is crucial for further progress.

\vspace*{3pt} 
\noindent \textbf{Acknowledgments.} This work was supported partially by the Major Research Plan of the National Natural Science Foundation of China under Grant No. 61991450 and the Shenzhen Key Laboratory of Marine IntelliSense and Computation under Contract ZDSYS20200811142605016.

{\small
\bibliographystyle{ieee_fullname}
\bibliography{egbib}
}

\end{document}